\documentclass{article}

\usepackage[final]{neurips_2024}

\usepackage{natbib}

\usepackage{color}
\usepackage{enumitem}
\usepackage{amsmath} 
\usepackage[utf8]{inputenc} 
\usepackage[T1]{fontenc}    
\usepackage{hyperref}       
\usepackage{url}            
\usepackage{booktabs}       
\usepackage{amsfonts}       
\usepackage{nicefrac}       
\usepackage{microtype}      
\usepackage{xcolor}         
\usepackage{graphicx} 
\usepackage{array} 
\usepackage{longtable}
\usepackage{tikz, tcolorbox}
\usepackage{courier}

\title{Fact or Fiction? Can LLMs be Reliable Annotators for Political Truths?}

\author{%
    Veronica Chatrath*, Marcelo Lotif, Shaina Raza\thanks{Denotes equal contribution.} \\
    Vector Institute for Artificial Intelligence \\
    Toronto, ON, M5G 0C6, Canada \\
    \texttt{\{veronica.chatrath, marcelo.lotif, shaina.raza\}@vectorinstitute.ai} \\
}

\begin{document}
\maketitle
\vspace{-5mm}
\begin{abstract}

Political misinformation poses significant challenges to democratic processes, shaping public opinion and trust in media. Manual fact-checking methods face issues of scalability and annotator bias, while machine learning models require large, costly labelled datasets. This study investigates the use of state-of-the-art large language models (LLMs) as reliable annotators for detecting political factuality in news articles. Using open-source LLMs, we create a politically diverse dataset, labelled for bias through LLM-generated annotations. These annotations are validated by human experts and further evaluated by LLM-based judges to assess the accuracy and reliability of the annotations. Our approach offers a scalable and robust alternative to traditional fact-checking, enhancing transparency and public trust in media.

\end{abstract}

\section{Introduction}
Political bias is the tendency to favour or oppose certain political ideologies, parties, or candidates in a way that skews impartiality, posing significant challenges in today's digital landscape \citep{bang2024measuringpoliticalbiaslarge}. While political bias refers to a large spectrum of topics like disinformation (false information spread intentionally to deceive) or misinformation (false information spread unknowingly) \citep{manualFactChecking}, this study is concentrated on distinguishing factual inaccuracies within politics. Traditional fact-checking approaches are not scalable given the vast amount of online information. Manual fact-checking methods, such as expert annotation and crowdsourcing, are viable alternatives but are often costly and time-consuming. Crowdsourcing lacks expert oversight and can lead to inconsistencies due to the varying expertise of contributors \citep{groupannotationbias}. These limitations highlight the need for efficient and scalable fact-checking approaches. 

\begin{figure}[ht]
    \centering
    \includegraphics[width=\linewidth]{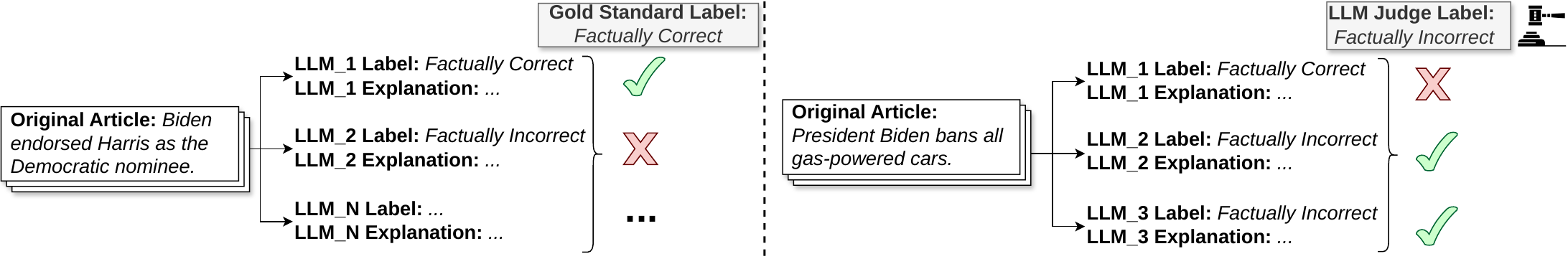}
    \vspace{-3mm}
    \caption{A working instance of our proposed LLM Annotator and Judge approach.}
    \label{fig:figure1}
\end{figure}

Recent studies have leveraged LLMs as annotators for various natural language processing (NLP) tasks including text classification \citep{gilardi2023chatgpt}, question answering \citep{ding2022gpt}, named entity recognition \citep{zhang-etal-2023-llmaaa,raza2024nbias}, and other related applications \citep{Choi2024MultiNewsCD, ni2024afacta, tan2024large, chatlabel}. Newer LLMs like GPT-4, Llama-3.1, Claude 3.5 Sonnet, and Mistral Large 2, updated for 2024, are equipped with extensive knowledge of the global political landscape and are often used for political analysis \citep{hernandes2024using}. Despite advancements, the use of these LLMs as annotators for labelling unstructured data and training machine learning models on political news remains an under explored application. Current annotation methods primarily rely on paid models such as GPT-4, leading to concerns regarding reproducibility, cost, and the quality of annotations.

We propose a scalable and cost-effective framework using open-source LLMs to reduce the effort and cost of annotating political data, as shown in Figure \ref{fig:figure1}. Our framework utilizes LLMs as both annotators and judges (evaluation models), making it adaptable to various LLMs. Our research questions include: \textit{Can LLMs effectively serve as annotators for political misinformation?}, \textit{Does including demonstrations (examples with prompts) improve LLM annotation performance?}, and \textit{How can we evaluate LLM-based annotations using LLMs as judges?}

Our key contributions are:
\begin{enumerate}[leftmargin=.5cm]
\item Construction of a high-quality, up-to-date human-annotated dataset.
\item Development of an LLM framework for binary annotation of factually correct and incorrect political news reporting. Our framework utilizes open-source LLMs for annotation, offering reproducible, accessible, and cost-free usage.
\item A reference-based comparison of the LLM annotations against gold (human verified) labels is presented. A human review process has been included to ensure quality and accuracy, and to address any discrepancies identified in the LLM annotation process.
\end{enumerate}

Empirical analysis indicates that annotations generated by LLMs closely match human annotations, as evidenced by high reference-based scores (Table \ref{tab:Method_accuracy}). Additionally, LLM-based evaluations for assessing label quality demonstrate strong performance of LLM annotators (Table \ref{tab:judge}).

\section{Related Works}

\textbf{Political Disinformation} 
Political disinformation involves framing or altering information to make a political position or candidate appear more attractive, or unattractive \citep{darkside}, and poses significant challenges, such as bias or toxicity \citep{raza2024unlocking}. In response, LLMs have emerged as a promising method to combat political disinformation by automating the detection and annotation of deceptive content, reducing the reliance on costly and time-consuming manual processes. Recent studies \citep{he2023annollm,alizadeh2024opensourcellmstextannotation,kim2024meganno+} highlight the potential of open-source LLMs in detecting and labelling disinformation. These models are shown to improve the efficiency and accuracy of disinformation detection, strengthening defences within the digital information ecosystem.

\textbf{LLM-as-an-Annotator}
Recent advances have demonstrated the potential of LLMs like OpenAI models \footnote{\url{https://openai.com/}}, such as GPT-3.5, GPT-4 and recently GPT-4o as effective annotators for various NLP tasks \citep{tan2024large}. These models can annotate data for classification and entity recognition through prompting methods. To maximize their utility, LLMs can be deployed within active learning loops \citep{zhang-etal-2023-llmaaa}, leveraging vast amounts of unlabelled data. Alignment tuning can further refine LLM annotations to match human preferences \citep{zhao2023survey}, mitigating potential biases.

Studies have shown that LLM-based annotations can match or exceed human performance across diverse domains, including social media analysis \citep{huang2023chatgpt}, computational science \citep{ziems2024can}, and medical information extraction \citep{goel2023llms}. Comparative analyses between LLMs and human annotators further highlight their potential \citep{gilardi2023chatgpt,he2023annollm,pavlovic-poesio-2024-effectiveness}. Building on these findings, our work employs LLMs as annotators to address political misinformation, an area yet to be fully explored in this context.

\textbf{LLM-as-a-Judge} Various LLMs such as GPT-3.5 \citep{gpt35} and GPT-4 \citep{gpt4} are increasingly utilized as evaluators, or judges, to ensure outputs align with human values \citep{zheng2024judging}. LLMs can assess the quality of model outputs against specific criteria such as accuracy, toxicity, and relevance \citep{dubois2023alpacafarm,zhou2023lima}, leveraging methods like asking for correctness or agreement with human annotations in a controlled experiment. In one LLM evaluation \footnote{{\href{https://arena.lmsys.org/}{LMSYS Chatbot Arena (lmsys.org)}}}, OpenAI's GPT-4o is at the top of the leaderboard, followed by Google's Gemini-1.5-Pro, and Meta's Llama-3.1-470b. Motivated by these results, we employ the newly available GPT-4o-mini \citep{gpt4omni} as a LLM judge in this work.

\section{Methodology}

\subsection{Dataset Construction}
We curated a dataset of news articles related to the North American political landscape, covering the period from May 6, 2023, to May 6, 2024. Data collection was conducted using automated scraping tools, including Newspaper and Selenium, with Google RSS as the main source. To ensure comprehensive coverage of electoral topics, a sample of 6,100 articles was selected based on the 26 topics and 16 sources listed in Appendix \ref{App:newssources}.

\begin{figure}
    \centering
    \includegraphics[width=.9\linewidth]{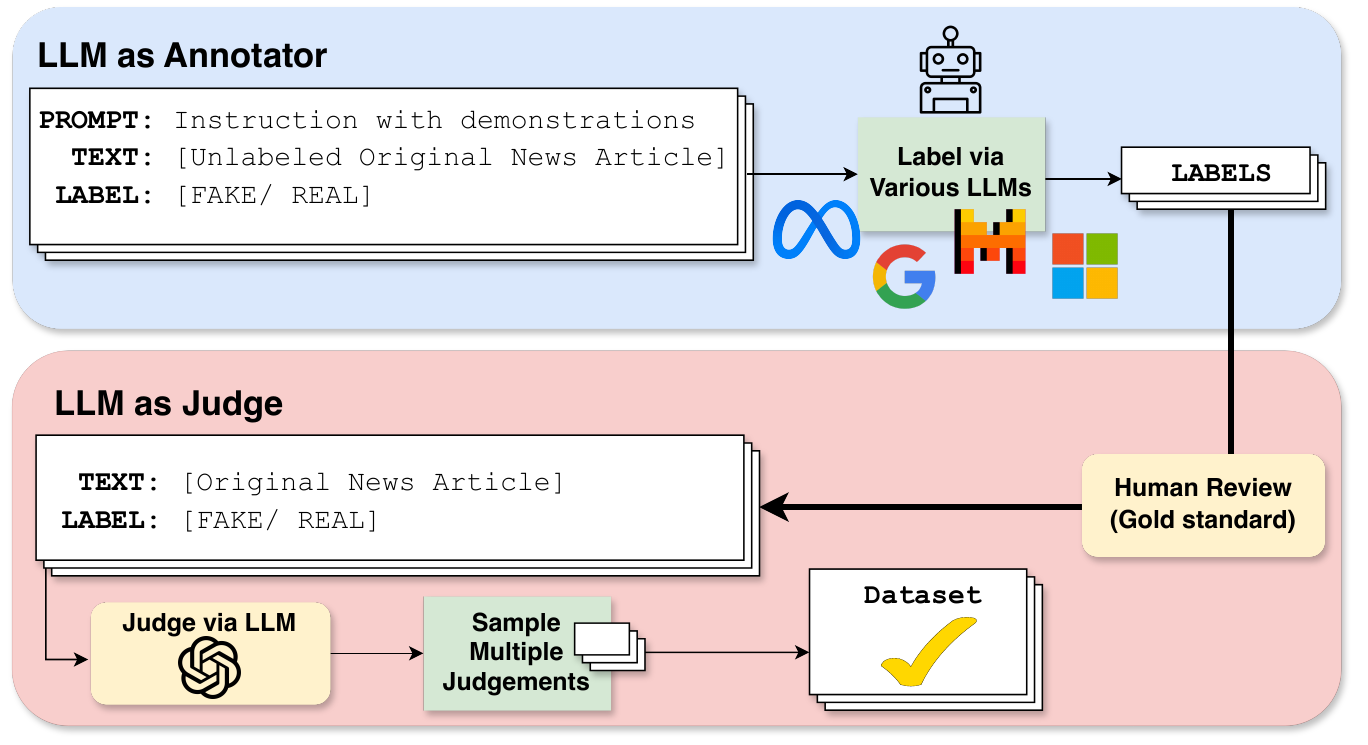}
    \vspace{-3mm}
    \caption{A visualization of our annotator (top) and judge (bottom) pipeline.}
    \label{fig:figure2}
\end{figure}

\subsection{LLM-as-an-Annotator}
We employ a suite of open-source LLMs, recognized for their robustness and accuracy in state-of-the-art applications \citep{Li_AlpacaEval_An_Automatic_2023}. These models include Llama-3-8B-Instruct \citep{touvron2023llama}, Llama-3.1-8B-
Instruct, Mistral-7B-Instruct-v0.3 \citep{jiang2023mistral}, Gemma-2-9b-Instruct \citep{gemma_2024}, and Phi-3-medium-128k-Instruct \citep{phi3}. 
The annotation task involves labeling each news summary article as either \textit{factually correct} or \textit{factually incorrect}. The LLMs are employed in both zero-shot and five-shot settings. The following is a snippet of the prompt used for annotation, with the entire prompt listed in Appendix \ref{App:prompt}.

\begin{tcolorbox}[title= Prompt used for Annotation, colframe=gray!50!white, halign=flush left, parbox=false]

\small

\textbf{role:} \texttt{user}, \textbf{content:} \texttt{You are a helpful news fact-checking bot trained to assess the accuracy of information. Analyze the given article and provide your assessment in the following format: ...}

\texttt{\textbf{Classification:} [Factually Correct/Factually Incorrect]}

\texttt{\textbf{Explanation:} Provide a concise, evidence-based explanation for your classification. Reference specific examples from the article and contradicting evidence from trusted sources, if applicable     ...}

\texttt{\textbf{Article to analyze:}} \texttt{article}

\end{tcolorbox}
The LLM-based annotations, determined by LLM majority voting, were reviewed by three human reviewers. The reviewers assessed each piece of data based on majority voting, resolving discrepancies, with unresolved cases escalated to a senior reviewer for the final decision. In this work, \textit{gold data} refers to the dataset portion annotated by an LLM and, subsequently, reviewed by humans.

\subsection{LLM-as-a-Judge}
We employ LLMs as evaluators (judges) to ensure objective and scalable assessment of our dataset's annotations. Inspired by previous works \citep{verga2024replacingjudgesjuriesevaluating, gptoutperform, Ding2022IsGA}, the evaluation task requires an LLM judge to assess the LLM annotator's response in a binary fashion (yes/no), determining whether it agrees with the annotation. The prompt used for the LLM-as-a-Judge experiments can be seen in Appendix \ref{App:judgePrompt}.

GPT-4o-mini is employed as the first judge, as it is quite affordable (\$0.150 per 1M input tokens, \$0.600 per 1M output tokens \footnote{{\href{https://openai.com/api/pricing/}{GPT-4o-mini Pricing} }}). To abide by our commitment to open-source science, Llama-3.1-70B-instruct is also used as a judge. Both models are employed to prevent self-enhancement bias \citep{zheng2024judging} which occurs when a model from the same family serves as both annotator and judge.

\section{Experiments}
\subsection{Experimental Setup}
\paragraph{Settings} Inference was run on a single A40 GPU with 4 CPU cores. For 6,000 samples across 5 LLMs, the process took about 16.67 hours, consuming 9.34 kWh and emitting 4.2 kgCO2e \citep{dodge_measuring_2022}. This carbon footprint is relatively low compared to fine-tuning. Open-source versions of Llama, Mistral, Gemma, and Phi models were used. GPT-4o-mini was employed as a judge via the OpenAI API, with total approximate experiment costs of \$2 USD on sample data.

\paragraph{Evaluation Metrics} Two approaches are used to evaluate the annotations: 1) Reference-based evaluation involves employing standard classification metrics by comparing the precision, recall, and accuracy of the LLM-generated annotations with the gold-standard ground truth labels, inspired by similar works \citep{gilardi2023chatgpt}. 2) LLM-as-a-judge-based evaluations, inspired by \citep{zheng2024judging}, are calculated as the percentage of times the LLM judge's assessment aligns with the LLM-based annotations, also known as Agreement Rate. This experiment is performed on 500 samples.

\begin{table}[ht]
    \centering
    \renewcommand{\arraystretch}{1.2}
    \begin{tabular}{p{5.2cm}p{2cm}p{1.5cm}p{1.5cm}p{1.5cm}p{2cm}}
    \toprule
     \textbf{Method} & \textbf{Acc.}  & \textbf{Prec.} & \textbf{Rec.} & \textbf{F1} \\ \midrule
     Llama-3-8B-Instruct (zero-shot)& 80.4\% &  82.3\% & 80.4\% & 80.7\% \\ \hline
     Llama-3-8B-Instruct (5-shot) & \textbf{89.3\%} & \textbf{89.3\%} & \textbf{89.3\%} & \textbf{89.2\%} \\ \hline
    Llama-3.1-8B-Instruct (zero-shot) & 74.5\% & 81.0\% & 74.5\% & 74.7\% \\ \hline 
    Llama-3.1-8B-Instruct (5-shot)& 82.1\% & 86.4\% & 82.1\% & 82.3\%  \\ \hline
    Mistral-7B-Instruct-v0.3 (zero-shot) & 80.0\% & 79.7\% & 80.0\% & 79.7\%  \\ \hline 
    Mistral-7B-Instruct-v0.3 (5-shot) & 82.5\% & 83.1\% & 82.5\% & 81.8\%   \\ \hline
    Gemma-2-9b-Instruct (zero-shot) & 77.6\% & 82.0\% & 77.6\% & 77.9\%   \\ \hline 
    Gemma-2-9b-Instruct (5-shot) & 83.3\% & 85.4\% & 83.3\% & 83.6\%  \\ \hline
    Phi-3-med-128k-Instruct (zero-shot) & 80.7\% & 80.8\% & 80.7\% & 80.7\% \\ \hline 
    Phi-3-med-128k-Instruct (5-shot) & 86.9\% & 87.1\% & 86.9\% & 86.7\%   \\ 
      \bottomrule

    \end{tabular}
    \vspace{2mm}
    \caption{Comparison of different LLM annotation's accuracy (Acc.), precision (Prec.), recall (Rec.), and F1-score (F1) against the gold labels in zero-shot and 5-shot settings. The best results are in \textbf{bold}.}
    \label{tab:Method_accuracy}
    \vspace{-5mm}
\end{table}


\subsection{Comparison with Gold Data}
We compare various LLMs as annotators against the gold data. This data was formed by a review team consisting of twelve volunteer members representing a diverse range of disciplines and demographics. Additional information on the composition of the gold data can be found in Appendix \ref{app:golddata}. Table \ref{tab:Method_accuracy} shows the performance of the LLMs in an annotator role. We observe that Llama-3-8B-Instruct (5-shot) outperforms all other models. Additionally, five-shot experiments outperform all zero-shot experiments, answering our second research question. The inference times for each run can be seen in Appendix \ref{app:inferencetime}.


\subsection{LLM-based Evaluation Results}

\begin{table}[ht]
    \centering
    \renewcommand{\arraystretch}{1.2} 
    \begin{tabular}{>{\centering\arraybackslash}p{4cm} >{\centering\arraybackslash}p{2.8cm} >{\centering\arraybackslash}p{2.8cm} >{\centering\arraybackslash}p{2.5cm}}
    \toprule
     \textbf{Method} & \textbf{AR    \ (GPT-4o-mini)} & \textbf{AR (Llama-3.1-70B)} & \textbf{AR    \   
  \  \ \ \ \ \ \ \ (Ground Truth)} \\ \midrule
     Llama-3-8B-Instruct & 76.4\% & \textbf{79.6\%} & 87.20\% \\ 
     Llama-3.1-8B-Instruct & 70.8\% & \textbf{73.2\%} & 78.0\% \\ 
     Mistral-7B-Instruct-v0.3 & \textbf{74.2\%} & 71.4\% & 81.4\%\\ 
     Gemma-2-9b-Instruct & 77.2\% & \textbf{77.6\%} & 84.0\%\\ 
     Phi-3-med-128k-Instruct & \textbf{74.4\%} & 74.0\% & 84.0\%\\
    \bottomrule
    \end{tabular}
    \vspace{1mm}
    \caption{Comparison of annotation agreement rates (AR) among LLMs, using GPT-4o-mini and Llama-3.1-70B as evaluators, against the ground truth agreement rate. Best judge results are in \textbf{bold}.}
    \label{tab:judge}
\end{table}

Table \ref{tab:judge} shows the annotation agreement rates (AR) for various LLM annotators, evaluated by GPT-4o-mini and Llama-3.1-70B against ground truth scores. Llama-3-8B-Instruct achieves the highest agreement rates, with 76.4\% for GPT-4o-mini and 79.6\% for Llama-3.1-70B, closest to the ground truth score of 87.2\%. Gemma-2-9b-Instruct also performs well, with consistent rates of 77.2\% and 77.6\% for the two judges, aligning relatively closely with its ground truth score of 84.0\%.

Notably, there are differences in how the two judges evaluate certain annotators. Mistral-7B-Instruct-v0.3 has a higher agreement rate with GPT-4o-mini (74.2\%) compared to Llama-3.1-70B (71.4\%), yet both are below its ground truth of 81.4\%. Phi-3-medium-128k-Instruct shows similar agreement with both judges (around 74\%) but falls short of its ground truth (84.0\%). 

The differences in agreement rates between the judges highlight the importance of using multiple evaluation methods for assessing LLM performance. Judges may have different preferences or criteria, leading to varying assessments. While a ground truth standard is useful, these discrepancies highlight the need for diverse perspectives to fully understand LLM performance.



\section{Discussion}
\textbf{Practical Impact} Our framework significantly improves the labeling process in NLP. By using LLMs for initial annotation, incorporating human oversight, and leveraging stronger LLMs for evaluation, we achieve a balanced solution that combines AI efficiency with human accuracy. This approach reduces costs and time compared to fully manual labeling, while maintaining high quality. It is practical for real-world applications like labeling news articles, analyzing social media, and processing customer feedback. For elections, this method can help analyze political content, fact-check claims, and track misinformation trends at scale.

\paragraph{Limitations and Future Work}

Our findings demonstrate LLMs' effectiveness as annotators, though potential biases in the annotation remain a concern, warranting further investigation \citep{das2024investigating}. Research shows LLM judges exhibit their own biases \citep{chen2024mllm}, impacting assessment reliability. They often favor their own responses \citep{zhao2021calibrate} and are susceptible to order and length biases \citep{li2024leveraging}. Prompt adjustments can alter LLM judgments, necessitating the use of multiple judges, including humans and various LLM models.

Despite controlled temperatures and carefully designed prompts, annotation responses and LLM judgments exhibit inherent stochasticity. These models rely on internal confidence mechanisms rather than explicit probability scores \citep{yan2024evaluating}. Potential mitigation methods include majority voting or aggregating multiple LLM responses. Future work will focus on incorporating associated images collected during initial data scraping, with the aim of developing a multi-modal model that leverages both textual and visual information for more robust analysis. 

\section{Conclusion}

In conclusion, this study highlights the potential of open-source LLMs as effective annotators for political misinformation detection, demonstrating their ability to produce annotations that align closely with human judgments. The variations in evaluation between different LLM judges highlight the need for multiple evaluation methods to comprehensively assess performance. While LLMs provide a promising and cost-effective approach for annotation, integrating them with human oversight enhances the reliability of the results. This framework has significant implications for improving labeling processes in NLP applications, especially in analyzing political content for sentiment analysis and misinformation detection. Future work should explore multi-modal approaches that incorporate visual data to further enhance political misinformation annotation and combat misinformation effectively.

\section{Social Impacts Statement}

Bias is inherently subjective, and its definitions widely vary across different disciplines. While our work addresses misinformation for news media, it is not intended to cover the full range of biases that exist globally. We have made possible efforts to mitigate bias, but acknowledge that our approach may have its limitations. Additionally, while our methods aim to identify bias and ensure accurate fact-checking, it is important that the data and techniques we release are used responsibly and ethically. The same data, if misused, could unintentionally fuel the spread of hate, which is not our intention. We have employed LLMs as annotators and evaluators, with human oversight included in the process. However, there is still a possibility of inherent biases in LLMs, and we recommend cautious use of both the data and methods developed. Lastly, while we focus on addressing misinformation that affects certain groups, our intent is not to exacerbate the sentiments of those already impacted by systematic inequalities. 

\newpage
\bibliography{neurips_2024}
\bibliographystyle{plainnat}

\newpage
\appendix

\section{Appendix}

\subsection{Prompt for Annotation}
\label{App:prompt}
The full prompt used for the annotation phase, as described in Section 3.2.

\begin{tcolorbox}[title= Prompt used for Annotation, colframe=gray!50!white, halign=flush left, parbox=false]

\textbf{role:} \texttt{user}, 

\textbf{content:} \texttt{You are a helpful news fact-checking bot trained to assess the accuracy of information. Your task is to analyze the given article and determine whether it is \textit{Factually Correct} or \textit{Factually Incorrect}.} 

\texttt{Fact-checking is the methodical process of verifying claims in public discourse or media reports. It is vital for countering misinformation and disinformation, thereby enhancing public knowledge and trust. Consider the following in your evaluation:} 

\texttt{\textbf{Misinformation:} Incorrect or misleading information shared without intent to harm.}

\texttt{\textbf{Disinformation:} Information that is knowingly false, often prejudiced, and disseminated with the intent to mislead.}

\texttt{Your analysis should include:} 

\texttt{Verification of key claims against multiple reliable sources. Identification of logical fallacies or statements that may mislead readers. Assessment of the context in which the information was presented, including the source’s history and potential motivations. Evaluation for any presence of hate speech, linguistic harm, or intent to spread prejudice.}

\texttt{Provide your assessment in the following format:} 

\texttt{\textbf{Classification:} [Factually Correct/Factually Incorrect]}

\texttt{\textbf{Explanation:} Provide a concise, evidence-based explanation for your classification. Reference specific examples from the article and contradicting evidence from trusted sources, if applicable.}

\texttt{Ensure to remain objective, basing your assessment strictly on facts and evidence rather than opinions or biases.}

\texttt{\textbf{Article to analyze:} {article}}

\end{tcolorbox}


\newpage
\subsection{LLM-as-a-Judge Prompt}
\label{App:judgePrompt}
\begin{tcolorbox}[title= Prompt used by the LLM Judge, colframe=gray!50!white, halign=flush left, parbox=false]

\textbf{Prompt:} \texttt{Determine which annotation model better aligns with the given label and text.}

\textbf{Text:} \texttt{[Insert Text Here]} \\
\textbf{Label:} \texttt{[Insert Label Here]} 

\textbf{Annotation from all Models:} \texttt{[Insert annotation from Model A, B, C, etc. here]} 

\textbf{Evaluation Criteria:} \\
\texttt{- Alignment Assessment: Evaluate which model's annotation aligns better with both the text and the label. \\
- Instruction: Indicate whether Model A, B, C, etc. provides an annotation that better fits the given label and text context.}

\texttt{Provide your evaluation and a brief justification.}

\end{tcolorbox}




\newpage

\subsection{Dataset News Sources and Topics}
\label{App:newssources}

\begin{table}[ht]
    \centering
    \begin{tabular}{>{\centering\arraybackslash}p{5cm} >{\centering\arraybackslash}p{5cm} >{\centering\arraybackslash}p{5cm}}
        \toprule
        \textbf{Topics} & \textbf{Candidates} & \textbf{Party Affiliations} \\
        \midrule
        \begin{itemize}
            \item Healthcare
            \item Climate Change
            \item Education
            \item Foreign Policy
            \item Tax Reforms
            \item Social Justice
            \item Racial Equality
            \item Gender Equality
            \item Economic Inequality
            \item Immigration
            \item Gun Control
            \item Culture-war
            \item Abortion
            \item Equality
            \item Democracy
            \item Environmental Policy
            \item Technology and Innovation
            \item Veterans Affairs
            \item Public Safety
            \item Mental Health
            \item Drug Policy
            \item Employment
            \item Trade Policies
            \item International Relations
            \item Judicial Appointments
            \item International Relation
        \end{itemize}
        &
        \begin{itemize}
            \item Presidential Candidates
            \item Congressional Elections
            \item House of Representatives
            \item State Governors
            \item State Legislature
            \item Local Government
            \item Mayor
            \item City Council
            \item County Officials
            \item Judicial
            \item Federal
            \item Prime Ministerial
            \item Parliamentary
            \item Premier
            \item Provincial Legislature
            \item Municipal
            \item Territorial
        \end{itemize}
        &
        \begin{itemize}
            \item Democratic
            \item Republican
            \item Independent
            \item Libertarian Party
            \item Green Party
            \item Constitution Party
            \item Liberal Party of Canada
            \item Conservative Party of Canada
            \item New Democratic Party
            \item Bloc Québécois
            \item Green Party of Canada
            \item People’s Party of Canada
            \item Tories
        \end{itemize} \\
        \bottomrule
    \end{tabular}
    \vspace{1mm}
    \caption{Topics, Candidates, and Party Affiliations scraped for the dataset.}
    \label{tab:topics_candidates_parties}
\end{table}
\vspace{5mm}
\begin{table}[ht]
    \centering
    \begin{tabular}{p{4cm} @{\hskip 1mm} p{4cm} @{\hskip 1mm} p{4cm} @{\hskip 1mm} p{4cm}}
        \begin{minipage}[t]{4cm}
            \raggedright
            \begin{itemize}
                \item \href{https://www.cnn.com}{CNN}
                \item \href{https://www.bbc.com}{BBC}
                \item \href{https://www.nytimes.com}{The New York Times}
                \item \href{https://www.theguardian.com}{The Guardian}
            \end{itemize}
        \end{minipage} &
        \begin{minipage}[t]{4cm}
            \raggedright
            \begin{itemize}
                \item \href{https://www.cbsnews.com}{CBS News}
                \item \href{https://abcnews.go.com}{ABC News}
                \item \href{https://www.foxnews.com}{Fox News}
                \item \href{https://www.aljazeera.com}{Al Jazeera}
            \end{itemize}
        \end{minipage} &
        \begin{minipage}[t]{4cm}
            \raggedright
            \begin{itemize}
                \item \href{https://www.reuters.com}{Reuters}
                \item \href{https://www.apnews.com}{Associated Press}
                \item \href{https://www.bloomberg.com}{Bloomberg}
                \item \href{https://www.usatoday.com}{USA Today}
            \end{itemize}
        \end{minipage} &
        \begin{minipage}[t]{4cm}
            \raggedright
            \begin{itemize}
                \item \href{https://www.realclearpolitics.com}{Real Clear Politics}
                \item \href{https://www.pewresearch.org}{PEW Research}
                \item \href{https://www.cbc.ca}{CBC}
                \item \href{https://www.globalnews.ca}{Global News}
            \end{itemize}
        \end{minipage} \\
        \bottomrule
    \end{tabular}
    \vspace{1mm}
    \caption{News Sources scraped for the dataset.}
    \label{tab:news_sources}
\end{table}

\newpage

\subsection{Performance and Inference Time Comparison of LLM Annotation Methods}
\label{app:inferencetime}

\begin{table}[ht]
    \centering
    \renewcommand{\arraystretch}{1.2}
    \begin{tabular}{p{4.2cm}p{1.3cm}p{1.3cm}p{1.3cm}p{1.3cm}p{1.8cm}}
    \toprule
     \textbf{Method} & \textbf{Acc.}  & \textbf{Prec.} & \textbf{Rec.} & \textbf{F1}  & \textbf{Time}\\ \midrule
     Llama-3-8B (zero-shot)& 80.4\% &  82.3\% & 80.4\% & 80.7\%   & 3 hr. 23 min.\\ \hline
     Llama-3-8B (5-shot) & \textbf{89.3\%} & \textbf{89.3\%} & \textbf{89.3\%} & \textbf{89.2\%} & 5 hr. 5 min. \\ \hline
    Llama-3.1-8B (zero-shot) & 74.5\% & 81.0\% & 74.5\% & 74.7\%  & 3 hr. 58 min. \\ \hline 
    Llama-3.1-8B (5-shot)& 82.1\% & 86.4\% & 82.1\% & 82.3\% & 4 hr. 15 min. \\ \hline
    Mistral-7B (zero-shot) & 80.0\% & 79.7\% & 80.0\% & 79.7\%   & 3 hr. 10 min. \\ \hline 
    Mistral-7B (5-shot) & 82.5\% & 83.1\% & 82.5\% & 81.8\%   & 4 hr. 5 min. \\ \hline
    Gemma-2-9b (zero-shot) & 77.6\% & 82.0\% & 77.6\% & 77.9\%   & 3 hr. 5 min. \\ \hline 
    Gemma-2-9b (5-shot) & 83.3\% & 85.4\% & 83.3\% & 83.6\%   & 5 hr. 20 min. \\ \hline
    Phi-3-med (zero-shot) & 80.7\% & 80.8\% & 80.7\% & 80.7\%   & 2 hr. 40 min. \\ \hline 
    Phi-3-med (5-shot) & 86.9\% & 87.1\% & 86.9\% & 86.7\%   &  3 hr. 52 min.\\ 
      \bottomrule

    \end{tabular}
    \vspace{2mm}
    \caption{Comparison of different LLM annotation's accuracy (Acc.), precision (Prec.), recall (Rec.), and F1-score (F1) against the gold labels in zero-shot and 5-shot settings. The best results are in \textbf{bold}.}
    \label{tab:Method_time_appendix}
\end{table}

This table, an extension of Table \ref{tab:Method_accuracy} from the main paper, includes an additional column indicating the inference time for each model. To conserve space, the model names in Column 1 have been abbreviated. 

\subsection{Gold Data Composition Process}
\label{app:golddata}

Our gold data review team consists of twelve volunteer members representing a diverse range of disciplines and demographics. The team includes three undergraduate students, four master’s students, and two professionals with doctoral degrees, spanning the fields of computer science, media studies, and political science. We ensured balanced gender representation, with six female and six male members from diverse cultural and ethnic backgrounds. Two subject matter experts in computer science and linguistics led the annotation efforts. The team followed detailed guidelines for identifying disinformation, ensuring consistent criteria across all annotations. After completing the annotation process, a sample of annotations was evaluated using \href{https://labelstud.io/}{Label Studio}. Disagreements in labeling were resolved through consensus discussions, ensuring high agreement and reliability in the final annotations. 
\newpage

\end{document}